\documentclass{article}
\usepackage{spconf,amsmath,graphicx}

\usepackage{graphicx}
\usepackage{amsmath}
\usepackage{amssymb}
\usepackage{booktabs}

\usepackage{ascii}
\usepackage[utf8]{inputenc} 
\usepackage[T1]{fontenc}    
\usepackage{url}            
\usepackage{booktabs}       
\usepackage{amsfonts}       
\usepackage{nicefrac}       
\usepackage{microtype}      
\usepackage[dvipsnames, svgnames, x11names]{xcolor}
\usepackage{float}

\usepackage{ntheorem}

\usepackage{colortbl}
\usepackage{enumitem}
\usepackage{wrapfig}
\usepackage{amsmath}
\usepackage{amssymb}
\usepackage{graphicx}
\usepackage{bbding}
\usepackage{pifont}
\usepackage{wasysym}
\usepackage{mathrsfs}
\usepackage{multirow}
\usepackage{textcomp}
\usepackage{footnote}
\usepackage{soul}
\usepackage{threeparttable}
\usepackage[ruled,vlined]{algorithm2e}
\PassOptionsToPackage{numbers, compress, sort}{natbib}
\usepackage[small]{caption}
\usepackage{hyperref}

\usepackage{color}
\definecolor{light-gray}{gray}{0.5}
\definecolor{pretty-blue}{RGB}{0, 113, 188}
\definecolor{linecolor1}{gray}{.95} 
\definecolor{linecolor}{gray}{.895} 
\definecolor{cGreen}{RGB}{57,181,74} 

\usepackage[capitalize]{cleveref}
\crefname{section}{Sec.}{Secs.}
\Crefname{section}{Section}{Sections}
\Crefname{table}{Table}{Tables}
\crefname{table}{Tab.}{Tabs.}

\title{CORSD: Class-Oriented Relational Self Distillation}
\name{Muzhou Yu\textsuperscript{1}, 
Sia Huat Tan\textsuperscript{2}, 
Kailu Wu\textsuperscript{2}, 
Runpei Dong\textsuperscript{1}, 
Linfeng Zhang\textsuperscript{2}, 
Kaisheng Ma\textsuperscript{2*}\thanks{*Corresponding author$:$\href{mailto:kaisheng@mail.tsinghua.edu.cn}{kaisheng@mail.tsinghua.edu.cn}}}
\address{\textsuperscript{1} Xi'an Jiaotong University, 
\textsuperscript{2} Tsinghua University}
\begin{document}
%
\maketitle
\begin{abstract}
   Knowledge distillation conducts an effective model compression method while holding some limitations:
   (1) the feature based distillation methods only focus on distilling the feature map but are lack of transferring the relation of data examples; (2) the relational distillation methods are either limited to the handcrafted functions for relation extraction, such as $L_{2}$ norm, 
   or weak in inter- and intra- class relation modeling. 
   Besides, the feature divergence of heterogeneous teacher-student architectures may lead to inaccurate relational knowledge transferring.
   In this work, we propose a novel training framework named Class-Oriented Relational Self Distillation (CORSD) to address the limitations. The trainable relation networks are designed to extract relation of structured data input, and they enable the whole model to better classify samples by transferring the relational knowledge from the deepest layer of the model to shallow layers. 
   Besides, auxiliary classifiers are proposed to make relation networks capture class-oriented relation that benefits classification task. Experiments demonstrate that CORSD achieves remarkable improvements. 
   Compared to baseline, 3.8$\%$, 1.5$\%$ and 4.5$\%$ averaged accuracy boost can be observed on CIFAR100, ImageNet and CUB-200-2011, respectively.
\end{abstract}
%

\section{Introduction}

Driven by the deep neural network (DNN), computer vision~\cite{Alpher09,Alpher03} has developed at an unprecedented speed. The success of SOTA model usually depends on high computing and storage costs, and various techniques for model compression have been proposed, such as pruning~\cite{Authors04}, quantization~\cite{Alpher06} and knowledge distillation~\cite{Author08,Author10, TF-KD}, in which the inspiration of knowledge distillation comes from the knowledge transferred from teacher model to student model.
Moreover, self distillation (SD)~\cite{BornAgain} has been proposed to distill knowledge from deeper layers to shallower layers within one model. Due to its efficiency, many researches ~\cite{CS-KD, PS-KD} have enforced various criteria based on SD.

Recently, relation learning has become popular in deep learning~\cite{Alpher07,Author18}.
However, in knowledge distillation, most of the methods with or without relation learning, are faced of the following two issues:
(1) Previous methods without relation learning only focus on the alignment of output logits~\cite{Author10} or features~\cite{Author08} from teacher to student but ignore the transferring of relation of data examples,leading to low-efficiency distillation. 
(2) Lately, some works propose to improve the performance of knowledge distillation by transferring relation of data samples but are limited in handcrafted functions for relation extraction, such as $L_{2}$ norm~\cite{Alpher07} or inner product~\cite{Author18}. These naive relation extraction methods often fail to make full use of the relational knowledge, \textit{i.e.}, inter-class contrastiveness and intra-class similarity. Furthermore, direct transferring of knowledge from teacher to student of different architectures ignores the hierarchical feature divergence within models, which may lead to inaccurate relation transferring in feature space.
\begin{figure*}[!ht] 
	\centering  
	\includegraphics[width=0.95\linewidth]{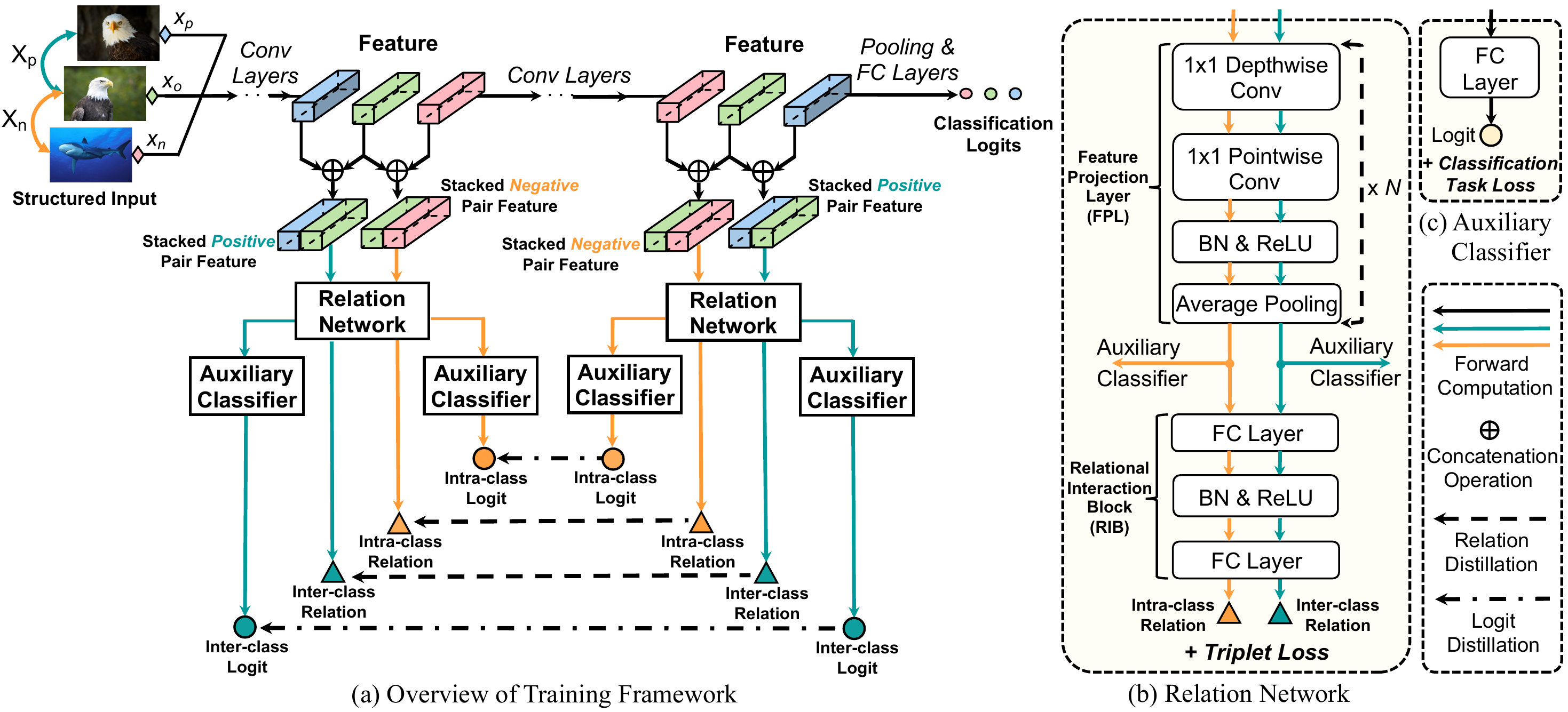}
	\caption{The overview of our proposed training framework Class-Oriented Relational Self Distillation (CORSD). \emph{Best viewed in color.}
	}
	\label{fig:2}
\end{figure*}

In this work, we propose a novel training framework named Class-Oriented Relational Self Distillation (CORSD) to overcome these obstacles. 
In order to make full use of sample relation, we use structured input where each input image is randomly coupled with a positive image of the same class and a negative image of a different class. Rather than using handcrafted functions for relation extraction, we design the trainable relation networks to extract the inter-class and intra-class relational knowledge of the structured input.
In order to train the relation networks for better relation modeling and reinforcement, we adopt the typical triplet loss~\cite{Author26} to further enhance the inter-class contrastiveness and intra-class similarity. As a result, the class-aware relation is obtained and enhanced, and then be transferred from the deepest layer of the backbone to shallow layers.
Moreover, auxiliary classifiers are proposed to help relation networks to capture class-oriented relation that is beneficial to classification task. 

To sum up, the contribution of this work can be summarized as follows: (1) We propose a novel training framework named CORSD. The trainable relation networks are designed to capture inter-class and intra-class relation of structured input, and the transferring of the relational knowledge from the deepest layer to shallow layers enables the whole model to better discriminate and classify input. (2) Auxiliary classifiers are utilized to help the relation networks for class-oriented relation extraction that are beneficial to classification task, leading to high-efficiency relational distillation. (3) Extensive experiments on three datasets across six models demonstrate that our proposed CORSD outperforms the SOTA distillation methods by a large margin. 

\section{Methodology}

\begin{table*}[t!]
\setlength{\abovecaptionskip}{0cm}
\setlength{\belowcaptionskip}{-0.2cm}
\setlength\tabcolsep{5pt}
\begin{center}
\resizebox{\textwidth}{!}{
\begin{tabular}{cccccccccccccc}
\toprule
\textbf{Model} & \textbf{Baseline} & \textbf{KD}~\cite{Author10} & \textbf{FitNet}~\cite{Author08} & \textbf{DML}~\cite{Alpher14} & \textbf{RKD}~\cite{Alpher07} &\textbf{AT}~\cite{Author09} & \textbf{SP}~\cite{Author18} & \textbf{CC}~\cite{CC} & \textbf{SD}~\cite{Author11} & \textbf{CS-KD}~\cite{CS-KD} & \textbf{TF-KD}~\cite{TF-KD} & \textbf{PS-KD}~\cite{PS-KD}  & \textbf{CORSD}\\
\midrule
ResNet18 & 77.09 & 78.34 & 78.57 & 78.72 & 78.53 & 78.45 & 79.02 & 78.76 & 78.64 & 79.51 & 78.43 & \underline{79.90}  & $\textbf{81.10}_{\color{cGreen}{~\uparrow\textbf{1.20}}}$\\

ResNet101 & 77.98 & 78.97 & 79.12 & 79.82 & 79.59 & 79.65 & 80.12 & 79.91 & 80.23& 81.12 & 80.04 & \underline{81.59}
& $\textbf{82.21}_{\color{cGreen}{~\uparrow\textbf{0.62}}}$\\

ResNeXt50-4 & 79.49 & 80.46&79.54&80.39& 80.94 & 81.05 & 80.67 & 80.72 & \underline{82.45}& 81.63 & 80.14 & 82.10 & $\textbf{82.50}_{\color{cGreen}{~\uparrow\textbf{0.05}}}$\\

WideResNet50-2 & 79.13 & 80.15 & 79.52 & 79.95& 80.54 & 80.42 & 80.78 & 80.94
&\underline{82.16}& 81.71 & 80.45 & 82.15 & $\textbf{83.02}_{\color{cGreen}{~\uparrow\textbf{0.86}}}$\\

WideResNet101-2&79.53
&80.52&79.65&80.69&80.46& 80.71 & 81.05 & 81.19 
&\underline{82.56}& 82.16 & 80.93 & 82.41 & $\textbf{83.55}_{\color{cGreen}{~\uparrow\textbf{0.99}}}$\\

SEResNet18&77.27
&78.43&78.49&78.58&78.17& 78.81 & 78.42 & 78.56 & 79.01& \underline{79.86} & 78.96 & 79.67 &  $\textbf{80.70}_{\color{cGreen}{~\uparrow\textbf{0.84}}}$\\

SEResNet50&77.69
&78.89&78.82&79.72&79.11 & 79.02 & 78.74 & 78.82 &\underline{80.56}& 80.01 & 79.23 & 80.12 &
$\textbf{82.54}_{\color{cGreen}{~\uparrow\textbf{1.98}}}$\\

PreactResNet18 & 76.05
&77.41&78.79&77.03&78.20 &78.01 & 78.77 & 78.34 
&78.12 & 78.87 & 78.01 & \underline{79.03}
& $\textbf{79.83}_{\color{cGreen}{~\uparrow\textbf{0.80}}}$\\

PreactResNet50&77.74
&78.26&79.12&78.48&79.15 & 79.15 & 79.57 & 79.59  
& \underline{80.12}& 79.77 & 78.96 & 80.10 & $\textbf{81.08}_{\color{cGreen}{~\uparrow\textbf{0.96}}}$\\

MobileNetV1 & 67.82
&67.55&\underline{71.78}&67.73&69.42& 71.33 & 71.57 & 70.78 
&71.39 & 71.45 & 70.67 & 71.23 
& $\textbf{71.90}_{\color{cGreen}{~\uparrow\textbf{0.12}}}$\\
\bottomrule
\end{tabular}
}
\end{center}
\caption{Top-1 accuracy (\%) on CIFAR100 
across a number of distillation methods. 
Boldface marks the best performing accuracy. The teacher model of all knowledge distillation methods is ResNeXt101-8, which the accuracy is 83.78 $\%$.
\label{Table:benchmark-table}}
\end{table*}
\begin{table}[t!]
\large
\setlength{\abovecaptionskip}{0cm}
\setlength{\belowcaptionskip}{-0.3cm}
\renewcommand\arraystretch{1.1}

\begin{center}
\resizebox{\columnwidth}{!}{
\begin{tabular}{cccccc}
\toprule
\textbf{Model} & \textbf{Baseline} & \textbf{SD}~\cite{Author11} & \textbf{CS-KD}~\cite{CS-KD} & \textbf{PS-KD}~\cite{PS-KD} & \textbf{CORSD}\\
\midrule

ResNet18 & 69.57 &
70.51 & 70.39 &\underline{70.59} & $\textbf{71.02}_{\color{cGreen}{~\uparrow\textbf{0.43}}}$\\
ResNeXt50-4 &77.62 & 78.47 & 78.35& \underline{78.76} 
& $\textbf{79.13}_{\color{cGreen}{~\uparrow\textbf{0.37}}}$\\ 
WideResNet50-2&78.47&\underline{79.02}& 78.76 & 78.91
& $\textbf{79.45}_{\color{cGreen}{~\uparrow\textbf{0.43}}}$\\


\bottomrule
\end{tabular}
}
\end{center}

\caption{Top-1 accuracy ($\%$) on ImageNet across several self distillation based methods. Boldface marks the best performing accuracy. \label{3}}
\end{table}
\begin{table}[h]
\setlength{\abovecaptionskip}{0cm}
\renewcommand\arraystretch{1}
\setlength\tabcolsep{15pt}
\begin{center}
\resizebox{\columnwidth}{!}
{
\begin{tabular}{cccc}
\toprule
\textbf{Model} & \textbf{Baseline} & \textbf{RKD}~\cite{Alpher07} & \textbf{CORSD}\\
\midrule
ResNet18 & 73.72&\textcolor{black}{75.78}& $\textbf{77.30}_{\color{cGreen}{~\uparrow\textbf{1.52}}}$\\

ResNet50&74.02&	\textcolor{black}{76.80}& $\textbf{78.33}_{\color{cGreen}{~\uparrow\textbf{1.53}}}$ \\



ResNeXt50-4&74.20&\textcolor{black}{77.32}& $\textbf{78.92}_{\color{cGreen}{~\uparrow\textbf{1.60}}}$\\

WideResNet50-2&74.57&\textcolor{black}{77.42}& $\textbf{78.97}_{\color{cGreen}{~\uparrow\textbf{1.55}}}$\\

SEResNet18&73.79& \textcolor{black}{77.17}& $\textbf{78.71}_{\color{cGreen}{~\uparrow\textbf{1.54}}}$\\

PreactResNet18&73.02&\textcolor{black}{75.61}& $\textbf{77.12}_{\color{cGreen}{~\uparrow\textbf{1.51}}}$\\
\bottomrule
\end{tabular}
}
\end{center}
\caption{Top-1 accuracy ($\%$) on CUB-200-2011. Boldface marks the best performing accuracy. The teacher model of RKD is ResNeXt101-8, which accuracy is 79.12$\%$.}\label{4}
\end{table}
For an input image $x_o$, we randomly choose another image with the same label as the positive image $x_p$ along with an image with any different label as the negative image $x_n$.
Then we make the positive pair $\mathbf{X}_p=(x_o,x_p)$ and the negative pair $\mathbf{X}_n=(x_o,x_n)$ together to be our structured input, which respectively encodes the intra-class and inter-class samples. Note that during every training epoch, each image in the training set is trained only once. Therefore, the batch size of training is a multiple of 3 and for fair comparison, the batch size of all other competing methods in section 3.1 is the same as ours. For dataset CIFAR, CUB-200-2011 and ImageNet, the batch size is 129, 129 and 255, respectively. As the above process of data split is completely random, the training setting brings no extra benefit to the results. The overview of CORSD is shown in Figure~\ref{fig:2}. Since all the relation networks, auxiliary classifiers are dropped during inference, it brings no extra computation or memory overheads.

\subsection{Relation Network Training}
\textbf{Relation Network Architecture} Assuming there are total $k$ convolution blocks in the backbone and larger index of convolution block represents deeper layer, the relation networks are attached to different depths of the backbone model. And the relation networks are independent along $k$ depths of convolution blocks and hold their own parameters. Specifically, our designed trainable relation networks at the $i_{\textit{th}}$ convolution block can be denoted as $\Phi$ parameterized by $\vartheta^{i}$. $\Phi$ is two-fold where the first part $g^{RN}$ is the Feature Projection Layer (FPL) consisting of several convolutional layers 
which serves for feature projection. And the second part is the learnable Relational Interaction Block (RIB) $\psi^{RN}$, comprising fully connected layers 
which are applied for final relation calculation. The detailed structure of relation network is shown in Figure~\ref{fig:2}(b).

\textbf{Relation Reinforcement and Distillation} The images $x_o$, $x_p$ and $x_n$ are fed into the backbone to obtain the corresponding features at different depths. Let $f_o^i$, $f_p^i$ and $f_n^i$ respectively be the features of $x_o$, $x_p$ and $x_n$ for the $i_{\textit{th}}$ convolution block. 
We first concatenate the structured input features in couples, \textit{i.e.},
$f_o^i$ is respectively concatenated with $f_p^i$ and $f_n^i$ for the
stacked positive and negative pairs, which can be described as $F_p^i=[f_o^i,f_p^i]$ and $F_n^i=[f_o^i,f_n^i]$. 
Then, the relation within the positive pair as well as the negative pair can be given as $R_p^i=\Phi(F_p^i;\vartheta^i)$ and $R_n^i=\Phi(F_n^i;\vartheta^i)$. $R_p^i$ and $R_n^i$ represent the distance measure within the positive pair and negative pair in feature space. Therefore, the smaller $R_p^i$ is, the closer relation within the positive pair is, and for $R_n^i$ vice versa.
Then we introduce the triplet loss $\mathcal{L}_{triplet}$ to train our relation networks to encourage closer interrelation within the positive pair (intra-class) and distant relation within the negative pair (inter-class) by margin $m$:
\begin{equation}
\setlength{\abovedisplayskip}{3pt}
\setlength{\belowdisplayskip}{3pt}
\mathcal{L}_{triplet}=\alpha\sum_{i=1}^{k}\max(R_p^i-R_n^i+m, 0),
\end{equation}
where $\alpha$ is the hyper-parameter and margin $m$ is usually set to 1. Then, the relation distillation loss $\mathcal{L}_{RD}$ is utilized to transfer the relation $R_p^i$ and $R_n^i$ from the deepest layer to shallow layers:
\begin{equation}
\setlength{\abovedisplayskip}{3pt}
\setlength{\belowdisplayskip}{3pt}
\begin{split}
     \mathcal{L}_{RD}=\beta\sum_{i=1}^{k-1}\big(\ell_2(R_p^i,R_p^{k})+\ell_2(R_n^i,R_n^{k})\big).
\end{split}
\end{equation}
where $\beta$ is the hyper-parameter and $\ell_2$ represents $L_2$ norm that induces the relation within the positive and negative pairs at shallow layers to approximate the corresponding relation at the deepest layer. Therefore, more knowledge of inter-class and intra-class information are distilled to shallow layers and thus the whole neural network learns more and better.

\subsection{Auxiliary Classifier Training}
 We propose auxiliary classifiers to help the relation networks further capture class-oriented relation that are beneficial to classification task. Our designed positive and negative auxiliary classifiers are both comprised of a single FC layer as the classification head. Similar to the relation networks, the auxiliary classifiers are depth-dependent, which can be parameterized with $\xi^i_p$ for the positive classifier and $\xi^i_n$ for the negative at the $i_{\textit{th}}$ block. Let $Z_p^i$ and $Z_n^i$ be the projected positive and negative pair features after FPL $g^{RN}$ in relation networks. Then, $Z_p^i$ and $Z_n^i$ are separately fed into the positive and negative auxiliary classifiers $\phi_p^{AC}$ and $\phi_n^{AC}$ for auxiliary classification task training. Because we train the features after FPL through auxiliary classifiers, we can get more class-oriented features. When these features are fed into RIB for relation extraction, they can help RIB capture more class-oriented relation, that is, more inter- and intra- class relation.
\begin{figure}[!t]
\setlength{\belowcaptionskip}{-0.3cm}
	\centering  
	\includegraphics[width=\linewidth]{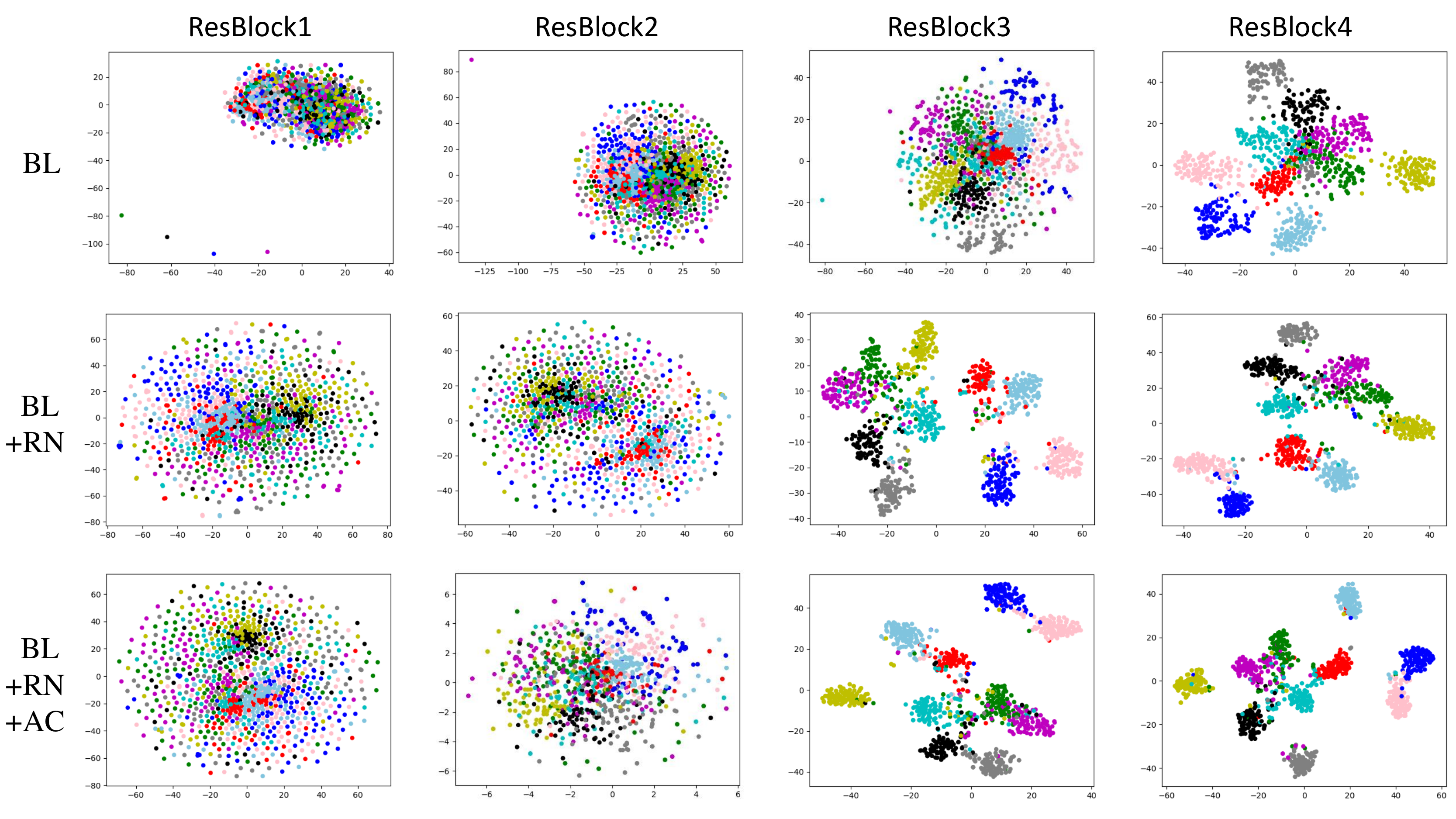}
	\caption{Visualization of feature distribution at different depths. BL represents baseline training, RN indicates relation network training and AC stands for auxiliary classifier training. \textit{Recommended zoom in for better view.}}
	\label{fig:3}
\end{figure}

\begin{figure*}[t!]\Large
\setlength{\belowcaptionskip}{-0.3cm}
\begin{center}
\begin{minipage}{0.27\linewidth}
\renewcommand\arraystretch{1}
\footnotesize
\setlength{\abovecaptionskip}{0cm}
\begin{center}
\resizebox{\columnwidth}{!}{




\begin{tabular}{c|cccccc}
\toprule
BL & \checkmark & \checkmark & \checkmark & \checkmark & \checkmark & \checkmark \\
\midrule
+RD & \ding{53} & \checkmark &  \ding{53} &  \ding{53} & \ding{53} & \ding{53} \\
+RN & \ding{53} & \ding{53} & \checkmark & \checkmark & \ding{53} & \checkmark \\
+AC & \ding{53} & \ding{53} & \ding{53} & \checkmark & \ding{53} & \checkmark \\
+LD & \ding{53} & \ding{53} & \ding{53} & \ding{53} & \checkmark & \checkmark \\
\midrule
Acc. &77.1 &77.9	&79.1	&81.0 &77.6& 81.1 \\
\bottomrule
\end{tabular}
}
\end{center}
\captionof{table}{Ablation study of Top-1 accuracy ($\%$) with ResNet18 on the CIFAR100 dataset.\label{6}}
\end{minipage}
\hspace{5pt}
\noindent
\begin{minipage}{0.7\linewidth}
\centering
\includegraphics[width=\linewidth]{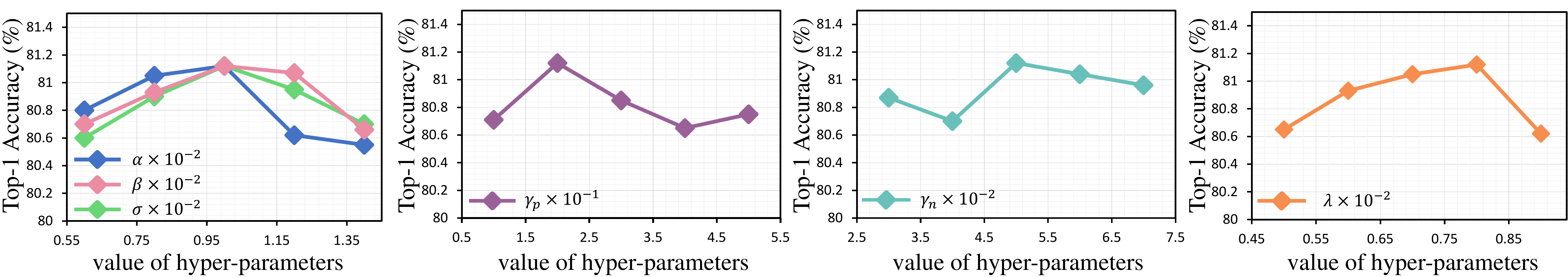}
	\caption{Sensitivity study of hyper-parameters with ResNet18 on the CIFAR100 dataset. \textit{Recommended zoom in for better view.}}
	\label{fig:hyper}
\end{minipage}
\end{center}
\end{figure*}





\textbf{Positive Auxiliary Classifier} For the positive projected feature $Z_{p}^{i}$, the positive auxiliary classifier loss $\mathcal{L}_{PAC}$ is composed of two components, the positive auxiliary task loss $\mathcal{L}_{PAT}$ and positive logit distillation loss $\mathcal{L}_{PLD}$. $\mathcal{L}_{PAT}$ is utilized to encourage $Z_{p}^{i}$ to perform the classification task, which can be formulated as:
\begin{equation}
\setlength{\abovedisplayskip}{3pt}
\setlength{\belowdisplayskip}{3pt}
    \mathcal{L}_{PAT}=\gamma_p(1-\lambda)\sum_{i=1}^{k}L_{CE}
    \big(\phi_p^{AC}(Z_p^i; \xi^i_p), y_{p}\big),
\end{equation}
where $\gamma_p$, $\lambda$ are hyper-parameters and $L_{CE}$ is the Cross Entropy and $y_{p}$ denotes the label of the positive pair $\mathbf{X}_p=(x_o,x_p)$ which is same as the label of image $x_o$ or $x_p$. Thus, we can get more class-oriented feature representation of $Z_{p}^{i}$.
Besides, the positive logit distillation loss $\mathcal{L}_{PLD}$ is designed to advance the training of the positive auxiliary classifier, which can be written as:
\begin{equation}
\setlength{\abovedisplayskip}{3pt}
\setlength{\belowdisplayskip}{3pt}
    \mathcal{L}_{PLD} = \gamma_p\lambda\sum_{i=1}^{k-1}D_{KL}
    \big(\phi_p^{AC}(Z_p^i;\xi^i_p),\phi_p^{AC}(Z_p^{k};\xi^k_p)\big).
\end{equation}
where $D_{KL}$ represents the KL divergence. 

\textbf{Negative Auxiliary Classifier} Similarly for $Z_n^i$, the negative auxiliary classifier loss $\mathcal{L}_{NAC}$ is also two-fold, \textit{i.e.}, the negative auxiliary task loss $\mathcal{L}_{NAT}$ and negative logit distillation loss $\mathcal{L}_{NLD}$. $\mathcal{L}_{NAT}$ is utilized to encourage the $Z_n^i$ to represent more discriminated inter-class features: 
$\mathcal{L}_{NAT}=\gamma_n(1-\lambda)\sum_{i=1}^{k}L_{CE}\big(\phi_n^{AC}
    (Z_n^i; \xi^i_n), y^*_{n}\big)$,
where $\gamma_n$ is the hyper-parameter and $y^*_{n}$ denotes a soft label of the negative pair $\mathbf{X}_n=(x_o,x_n)$. This soft label can be written as $y^*_{n} = 0.5 y_{o} + 0.5 y_{n}$, where $y_{o}$ and $y_{n}$ are the labels of image $x_o$ and $x_n$. Similarly, we design the negative logit distillation loss $\mathcal{L}_{NLD}$ to promote the training of the negative auxiliary classifier: 
$\mathcal{L}_{NLD} = \gamma_n\lambda\sum_{i=1}^{k-1}D_{KL}
    \big(\phi_n^{AC}(Z_n^i; \xi^i_n),\phi_n^{AC}(Z_n^{k}; \xi^k_n)\big)$.
Assisted by the negative auxiliary classifiers,
the relation networks can obtain more class-oriented
features, and then make use of the discriminated features
for better relation modeling, that is class-oriented relation.

Besides the feature's relation alignments, we propose logit distributions calibration to further reinforce inter-class and intra-class relation of the backbone logits, that is minimized intra-class relation and maximized inter-class relation: $\mathcal{L}_{logit} = 
    \sigma(\psi^{RN}(\tilde{\mathcal{F}}_p;\eta)-
    \psi^{RN}(\tilde{\mathcal{F}}_n;\eta))$
where $\sigma$ is the hyper-parameter, $\tilde{\mathcal{F}}_p$, $\tilde{\mathcal{F}}_n$ denotes the stacked positive/negative backbone logit.
\section{Experiments}
The experiments are conducted with six different kinds of models, including ResNet~\cite{Alpher09}, PreActResNet~\cite{Alpher10}, SENet~\cite{Alpher11}, ResNeXt~\cite{Alpher12}, WideResNet~\cite{Author20}, MobileNetV1~\cite{Author21},
and are evaluated on three popular datasets, including CIFAR~\cite{Author22}, ImageNet~\cite{Alpher13} and CUB-200-2011~\cite{Author23}. Eleven kinds of distillation methods are used for comparison, \textit{i.e.} KD~\cite{Author10}, SD~\cite{Author11}. 
The baseline represents the model trained from scratch without any distillation.
Across all models in Table~\ref{Table:benchmark-table}, the average training time and memory of CORSD is about 152$\%$ and 163$\%$ increasement to baseline. Take ResNet18 trained on CIFAR100 as example, the training time of CORSD and baseline is 4.3h and 2.6h, and the training memory of CORSD and baseline is 2.4GB and 1.4GB. Overall, the training overhead of CORSD is acceptable and can be handled by commercial GPU.

\textbf{Experimental Results} Table~\ref{Table:benchmark-table}
shows the test accuracy of CORSD and other distillation methods on CIFAR100. We can conclude that CORSD brings significant accuracy boost compared to baseline and other competing methods.
Specifically, CORSD surpasses SD by a large margin. On average, 1.09$\%$ higher accuracy can be observed on CIFAR100.
The classification accuracy on ImageNet is shown in Table~\ref{3}. 
CORSD reveals the superiority compared with other methods. 
The classification results on CUB-200-2011 is shown in Table~\ref{4}. 
CORSD leads to 1.54$\%$ superiority on average across all models than RKD, ranging from 1.51$\%$ for PreactResNet18 to 1.60$\%$ for ResNeXt50-4. In conclusion, our proposed CORSD achieves SOTA and achieves consistent and significant accuracy improvement on different neural networks and datasets compared with other distillation methods. 

\textbf{Visualization} In order to have a better understanding of the enhancement of inter-class and intra-class relation of CORSD, we visualize the distributions of samples in feature space at different depths.
As depicted in Figure~\ref{fig:3}, t-SNE visualizations~\cite{tSNE} with ResNeXt101-8 trained on CIFAR100 are conducted. 
When we successively add relation network training and auxiliary classifier training to the baseline, the performance of each block to cluster samples in feature space are much more remarkable, which can help the deep neural network to better classify samples.

\textbf{Ablation Study and Sensitivity Study} 
As shown in Table~\ref{6}, we can observe that relation network training (RN), auxiliary classifier training (AC) and logit distribution calibration (LD) in our method has its individual effectiveness and they can be utilized together to achieve better performance than baseline (BL) as well as relation self distillation (RD) using handcrafted function $L_{2}$ norm for relation extraction without our proposed relation network. 
The sensitivity study of each hyper-parameter is shown in Figure~\ref{fig:hyper}. Since the values of different losses vary in one order of magnitude at most, the values of hyper-parameters for different losses differ from $10^{-1}$ to $10^{-2}$ for the loss balancing. It can be observed that our method is robust to the choice of hyper-parameters and has consistent accuracy boost to other SOTA method PS-KD, which the accuracy is 79.9$\%$.
In this paper, we adopt the best setting ($\alpha=\beta=\sigma=0.01$, $\gamma_p=0.2$, $\gamma_n=0.05$, $\lambda=0.8$) for all experiments. 

\section{Conclusions}
In this work, we propose a novel training framework named CORSD for performance boosting of models on classification task. 
The trainable relation networks and auxiliary classifiers are designed to capture and reinforce the relation of structured input, taking full advantage of inter-class contrastiveness and intra-class similarity, which conspicuously bring benefits to classification performance.

{
\bibliographystyle{IEEEbib}
\bibliography{main}
}
\end{document}


\maketitle

\section{More Detailed Experimental Setting}
The experiments of image classification are conducted with six different kinds of convolutional neural networks, 
including ResNet~\cite{Alpher09}, PreActResNet~\cite{Alpher10}, SENet~\cite{Alpher11}, ResNeXt~\cite{Alpher12}, WideResNet~\cite{Author20}, MobileNetV1~\cite{Author21}.

For CIFAR and CUB-200-2011 experiments, we adopt random cropping, horizontal flipping and cut-out~\cite{Cutout} for data augmentation during training. All models are trained using SGD optimizer with momentum set to 0.9, 0.0005 weight decay, 0.1 initial learning rate, and we use one-cycle~\cite{OneCycle} learning rate scheduler as default. In addition, all models are trained for 300 epochs and the batch size is 129. Note that the structured input in TORSD is triplet, and thus the value of batch size should be a multiple of three. For ImageNet experiments, we follow the standard PyTorch practice but the training batch size for TORSD is 255.


All the experiments are conducted with PyTorch~\cite{Pytorch}, running on NVIDIA 2080Ti and Tesla V100 GPU devices.

\section{More Visualization and Explanation}
\begin{figure}[!htbp] 
	\centering  
	\includegraphics[width=\linewidth]{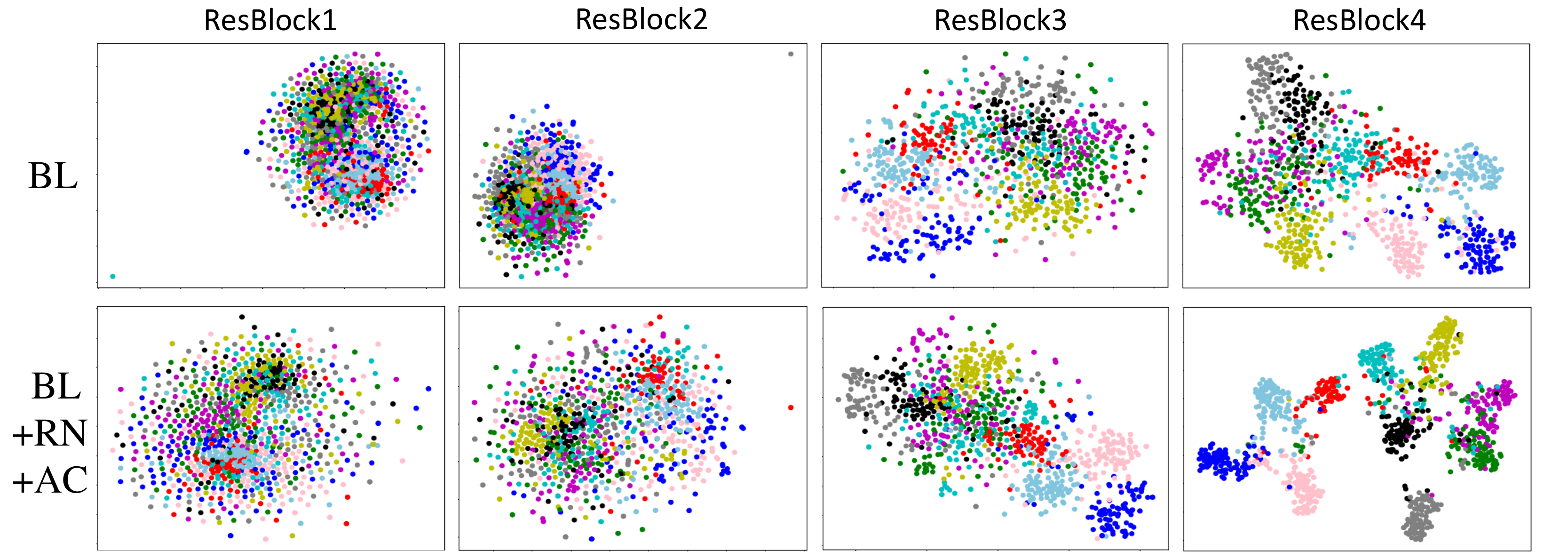}
	\caption{Visualization of feature distribution at different depths.
	BL represents baseline training, RN indicates relation network training and AC stands for auxiliary classifier training.}
	\label{fig:vis}
\end{figure}

\paragraph{Visualization} In Figure 3 of the submitted paper, we have shown the results with ResNeXt101-8 trained on CIFAR100, which visualize the distribution of samples in feature space at different depths. Due to the limitation of the paper length, here we conduct more t-SNE visualization~\cite{tSNE} with ResNet18 trained on CIFAR100. The results are shown in Figure~\ref{fig:vis}. As is consistent with the conclusion in the submitted paper, the ability of deep neural network at each depth to concentrate features is significantly improved with the proposed relation network training and auxiliary training. It indicates that transferring task-oriented relation of structured data from deepest layer to shallow layers can help the model obtain more separable feature distribution at different depths, which is beneficial to better classify data samples.

\paragraph{Sorted Separability} Similar to Table 5 in the submitted paper, we further use SSE/SSB to evaluate the sorted separability at each depth for ResNet18 trained on CIFAR100. A smaller SSE/SSB indicates the feature space is denser. Table~\ref{1} summarizes the results. It can be observed that with our proposed TORSD, the feature space is much denser than the baseline training. Furthermore, it is noteworthy that the more clustering feature space at different depths, the higher accuracy the model achieves.

\begin{table}[h!]
\begin{center}
\begin{tabular}{cccc}
\toprule
\textbf{Depth} & \textbf{BL} & \textbf{BL+RN+AC}\\
\midrule
ResBlock1 & 23.52 & 12.13\\
ResBlock2 & 15.87 & 7.68\\
ResBlock3 & 7.13  & 5.26\\
ResBlock4 & 5.06  & 2.74\\
\midrule
Acc. &77.09 & 80.80\\
\bottomrule
\end{tabular}
\end{center}
\caption{Measurement of sorted separability for each depth and accuracy ($\%$) with ResNet18 trained on CIFAR100. BL represents baseline training, RN indicates relation network training and AC stands for auxiliary classifier training.\label{1}}
\end{table}

\bibliographystyle{named}
\bibliography{main}